\documentclass[11pt,a4paper]{article}
\usepackage[hyperref]{naaclhlt2019}
\usepackage{times}
\usepackage{latexsym}
\usepackage{url}
\usepackage{times}
\usepackage{latexsym}
\usepackage{balance}       
\usepackage{graphics}      
\usepackage[T1]{fontenc}   
\usepackage{txfonts}
\usepackage{mathptmx}
\usepackage{color}
\usepackage{booktabs}
\usepackage{textcomp}
\usepackage{enumitem}
\usepackage{forest,adjustbox}
\usepackage{nameref}

\aclfinalcopy 

\title{Using Structured Representation and Data: A Hybrid Model for  Negation and Sentiment in Customer Service Conversations}

\author{Amita Misra, Mansurul Bhuiyan, Jalal Mahmud, and Saurabh Tripathy\\
  IBM-Research, Almaden \\
  San Jose, CA, USA \\
  {\tt amita.misra1|mansurul.bhuiyan|jumahmud|Saurabh.Tripathy2@ibm.com} \\}

\date{}

\begin{document}
\maketitle
\begin{abstract}

Twitter customer service interactions have recently emerged
as an effective platform to respond and engage with customers. In this work, we explore the role of negation in customer service
interactions, particularly applied to sentiment analysis.
 We define rules to identify true negation cues and scope more suited to conversational data than existing general review data. Using semantic knowledge and syntactic structure from constituency parse trees, we propose an algorithm for scope detection that performs comparable to state of the art BiLSTM. We further investigate the results of negation scope detection for the sentiment prediction task on customer service conversation data using both a traditional SVM  and a Neural Network. We propose an antonym dictionary based method for negation applied to a CNN-LSTM combination model for  sentiment analysis. Experimental results show that the antonym-based method  outperforms the previous  lexicon-based and neural network methods.
\end{abstract}

\section{Introduction}

Negation has been described as a polarity influencer \cite{Wilson2009} and therefore it has to be taken into consideration while designing a sentiment prediction system, but how important it is in twitter customer service conversations? For example, both the  customer service  tweets in Table~\ref{sent-ex1} have an explicit negation cue  but the effect of cue words on the polarity differ. The first tweet has a negation cue [\emph{don't}] that changes the positive polarity of the  words in the scope  [\emph{think you do understand}]. Additionally, tweet 1 has a hashtag [\emph{Misleading}] which could be a strong negative signal on its own. The second tweet has a cue word [\emph{not}] but it does not negate the words in that sentence or change their polarity. The negation cue [{\it not}] in the second tweet is not a true negation cue, and hence it has no scope.

\begin{table}[!htb]
	\begin{center}
			\begin{small}
			\begin{tabular} { p{0.4cm}  p{5.0cm}  p{1.5 cm}}  
				\toprule
				\bf {S.No}& \bf  {Tweet} & \bf {Sentiment}  \\ 
				1 & @Username I \emph{don't} think you do understand.  Buyers and Sellers deserve to know facts,User actively prevents accurate feedback. \#Misleading.& Negative\\ 
				2& @Username Sorry to hear this. Have you had a chance to call/chat us? If \emph{not}, we can look into options: & Positive \\ 
				 \\ 			
				\bottomrule
			\end{tabular}
				\end{small}
		\caption{\label{sent-ex1} Customer Service Conversation. }
	\end{center}
\end{table}

Negation can be expressed in different ways in natural language.
It may be through the use of explicit negation cues
such as {\it no, not and never} that have a morphologic indication
of a negative meaning. This also includes a group of broad
or semi negatives words (e.g. barely, hardly, and seldom) that
have a negative meaning but are without any morphological
negative.  This has been also referred to as clausal or syntactic negation \cite{Quirk1985,Givon1993}. These cue words are often used to
negate a statement or an assertion that expresses a judgment
or an opinion. However in some contexts, these cue words function as
exclamations, and not as true negation cues. These false cues  do not change the 
sentiment polarity of the following expression, and hence do not have any associated  scope.  We define
rules to identify true negation cues and their scopes more
suited to conversational data than existing general review
data. 

The impact of negation has been studied  in domains such as biomedical, literary texts, and  on-line  reviews \cite{BIOScopeSzarvas,Morante2008,Councill2010,Reitan2015,konstantinova2012review}; however, none of the previous corpora are  conversational in nature. Scope definitions may depend on the domain. \citet{Reitan2015} showed that negation scope algorithm trained on a twitter domain struggled when tested on a medical domain. Majority of the previous work in scope detection has been dominated by SVMs or Neural Networks, which require expensive annotated training data. Scope  annotation is costly and  time-intensive as all  the scope conflicts have to be resolved by mutual discussion amongst expert annotators. Our main motivation is to create a system that does not require a huge amount of training data for scope detection, but has  comparable performance to  machine learning models that require annotated training data. The proposed method  uses constituency parse trees and semantic knowledge to predict scope. 
 The  results in Table~\ref{tab-nsdresults} show that the method is  comparable to state of the art BiLSTM model from~\cite{Fancellu2016NeuralNF} on gold negation cues  for scope prediction. Since our method does not need expensive training data,  we could also use this method to predict on other negation data sets. However, our aim here was first to test if the predicted negation scope improves sentiment in conversations.
 
For a real time sentiment prediction system, we need both a cue prediction system to determine the true negation cues, and scope detection.  As a first step, we use a data based approach to train an SVM  to predict true negation cues.  It's much faster and simpler to get annotated data for cue prediction, a binary task as compared to scope detection, which is a sequence labeling task. 
This is followed by a second step of constituency tree-based negation scope detection for predicted cues. The last step applies negation prediction coupled with antonym dictionary to improve the sentiment performance for a  combination CNN-LSTM model.

 The contributions of this paper are:

\begin{itemize}
	\item{Negation scope rules more suited to conversational data. }
	\item A constituency-tree based approach for scope detection that uses both semantic and structural information, and does not require annotated data for scope.
	\item An antonym based negation applied to a combination CNN-LSTM model for sentiment prediction in conversations.\\
\end{itemize}

 We begin with a discussion of related work in Section~\ref{related-work}, followed by  negation corpus  in Section \ref{neg_data}, and negation cue and scope detection experiments in Section~\ref{neg_model}. Next, we show the effect of introducing negation detection for the sentiment task  in Section \ref{sentiment}. We then  compare and contrast the twitter conversational sentiment data to previous datasets in Section \ref{disc}. Finally, Section \ref{conc} presents the conclusions and future directions.

\section{Related Work} 
\label{related-work}
Initial studies on negation scope detection were performed in Biomedical domain including medical reports, biological abstracts, and papers \cite{BIOScopeSzarvas}. \citet{Morante2009} used a 2 step approach: first, a decision tree to predict negation cues, followed by a  CRF meta-learner to predict negation scope. The model used a combination of  k-nearest neighbors, a support vector machine, and a  CRF. The research in this field was further enhanced by a shared SemEval 2012   negation and scope resolution task \cite{SEM2012SharedMorante}. The organizers  released a cue and scope annotated corpus of Conan Doyle stories. 

\citet{Read2012UIO1} described both, a rule-based and a data driven approach  for scope resolution. Both the methods were driven by the hypothesis that syntactic units correspond to scope annotations. The rule-based approach  used heuristic rules based on POS tags and constituent category labels, while machine learning used SVM based ranking of syntactic constituents.
Limited rule based system obtained similar results to the data-driven system on a held-out set. This result was particularly note-worthy since getting sufficient scope annotation training data for every new domain is quite expensive, and requires trained annotators. A comparison of these results motivated us to further develop the rule-based system for the conversational domain using both semantic information and syntactic structure.
\cite{Councill2010,LapponiUiO2,Reitan2015} used CRF-based sequence labeling using features from dependency tree. \citet{packard14} used hand-crafted heuristics to traverse Minimal Recursion Semantics (MRS). However, if a reliable representation for a sentence could not be created, their system used a fall back mechanism based on \citet{Read2012UIO1}. \citet{Fancellu2016NeuralNF} showed that a neural network based model using a BiLSTM outperformed the  previously developed classifiers on both scope token recognition and exact scope matching for in domain testing but not on a different domain. The authors noted that when tested on a different test set from  Wikipedia, \citet{whiteS12}'s model built on constituency-based features performed better.

A survey on the role of negation in sentiment analysis was done by \cite{Wiegand2010} stating that negation expressions are ambiguous i.e. in some contexts do not function as a negation and, therefore, need to be disambiguated. Rules of composition were defined by \citet{Moilanen2007} on the syntactic representation of a sentence to account for negation and the modeling paradigm could be  applied to determine the sub-sentential polarity of the sentiment expressed. \cite{Councill2010} showed that a CRF based negation enhanced classifier improved the F-score of positive on-line reviews by 29.5\% and 11.4\% for negative. Much recent progress in the field has been in connection with the 
"The International Workshop on Semantic Evaluation" (SemEval) \cite{nakov2013eveval}.  Since 2013 the workshop has included shared tasks on "Sentiment Analysis in Twitter". Most of the top performing systems submitted used just a simple punctuation model that assigns a negation cue scope over all the terms to the next punctuation \cite{Tangsem14,Miura2014,Mohammadsem13}.
\cite{Kiritchenko2014} reported an improvement of up to 6.5 percentage points when handling negated context on the SemEval-2013 test set. Using the simple punctuation model for scope detection, an improvement of upto 6\% was reported  by  \cite{Reitan2015}.

With the recent advances in deep learning and use of embeddings, the CNN and  LSTM based models have shown to outperform traditional SVM and lexicon based methods for sentiment in twitter and review domain.  \citet{Kim14} applied  a CNN based model to numerous document classification tasks, and  improved the sentiment state of the art using a CNN architecture with one layer of convolution trained using word vectors obtained from \citet{Mikolov13} on 100 billion words of Google News. \citet{YinS15} combined  different word embeddings using multichannel CNN. \citet{WangJL16} showed that a combination CNN-LSTM outperformed  CNN for sentiment task. \citet{ShinLC17}  integrated sentiment embeddings in a  CNN to build simpler  high-performing models with much smaller word embeddings. 
However, none of the previous work has explored negation coupled  with antonyms to get a better sentence representation for sentiment prediction. 

\section{Conversational Negation Corpus }
\label{neg_data}

\begin{table}[!htb]
	\begin{center}
		\begin{small}
			
			\begin{tabular} {  p{1.0 cm}  p{1.0 cm} p{1.0 cm}  p{1.0 cm}   p{1.0 cm}}  
				\midrule
				hardly   & lack     & lacking & lacks   & neither \\ 
				no       & nobody   & none    & nothing & nowhere \\ 
				cant     & arent    & dont    & doesnt  & didnt   \\ 
				havent   & isnt     & mightnt & mustnt  & neednt  \\ 
				shouldnt & wasnt    & werent  & wouldnt & without \\ 
				seldom   & scarcely & wont    & never   & aint    \\ 
				barely   & nor      & not     & hadnt   & rather \\ 
				hasnt    & shant    &         &         &        \\  
				\midrule
			\end{tabular}
		\end{small}
		\caption{\label{cue-lexicon}  Negation cue lexicon. }
	\end{center}
\end{table}


We selected conversations from the Twitter customer service pages of different companies and downloaded 89552 customer service tweets in total\footnote{ Comapny names are anonymous for annotation}. A lexicon of explicit cue words that may act as indicators of negation was primarily adopted from \cite{Councill2010,Reitan2015}. It was further extended to include  semi negative words. The final set of cues used is shown in Table~\ref{cue-lexicon}. We then extracted 23243 tweets  containing explicit negation cues giving a frequency of 26\%. In contrast, the equivalent numbers for BioScope corpus \cite{BIOScopeSzarvas}  and for Twitter corpus \cite{Reitan2015} are 13.8 \% and 13.5\%  respectively. \citet{tottie1991} presented a comprehensive taxonomy
of  English negations and stated that frequency of negation is 12.8\% in written English. In  another statistical study on negation, \cite{biber1999} reported that negation is much more frequent in conversation as compared to written discourse.  Since we had a lot more negation cue occurrences, we divided the tweets into 5 different groups based on the number of negation cues present in each tweet. A random  sample was selected from each group based on the number of instances in each group giving a dataset of 2000 tweets.  A separate set of 100 tweets was used as a development set to help formulate the rules and study negation patterns.  Every tweet was annotated by a pair of annotators. To test the robustness of guidelines, we measured  inter-annotator agreements (IAA) for each pair of raters using the token level and full scope measures as used in previous work \cite{Reitan2015}. The token level is the percentage of tokens annotators agreed upon. Since the average number of tokens in scope is far less than the number outside the scope, this is a skewed measure.  For full scope, it is the percentage of scopes that have a complete and exact match amongst annotators (PCS). After an initial annotation phase of 1000 tweets, the average token level agreement was 0.95  and full scope was  0.78.  All the scope conflicts were mutually resolved after discussion. Corpus statistics are shown in Table \ref{tweet-dist}.
The average number of tokens per tweet is 22.3, per sentence is 13.6 and average scope length is 2.9.
\begin{table}[!htb]
	\begin{center}
		
		\begin{small}
			
			\begin{tabular} { p{4cm}  p{2cm}   }  
				
				\midrule
				Total negation cues & 2921\\
				True negation cues & 2674 \\
				False negation cues & 247 \\
				Average scope length & 2.9 \\
				Average sentence length& 13.6 \\ 
				Average tweet length& 22.3  \\ 
				\bottomrule		
			\end{tabular}
			

		\end{small}
		
		\caption{\label{tweet-dist} Cue and token distribution in the conversational negation corpus. }
		
	\end{center}
\end{table}
\subsection{Annotation Guidelines}
We define rules to identify true negation cues and their scopes more suited to twitter customer service conversational data than existing general review data, which has its own characteristics such as brevity and skewed distribution towards negative polarity (Sec. \ref{disc}). 
The guidelines described here were adapted from \citet{Councill2010} but modified for customer service conversations. 
Nouns and adjectives are key indicators of sentiment \cite{Hu2004,Pang2008} and hence  we had a more restricted scope for noun and adjectives as compared to verbs and adverbs. In the following examples, the  cue is underlined and the scope  is marked in bold.
\begin{itemize}
	\item  Annotating the negation cue.
	\begin{itemize}
		\item False Negation: Some negation cues can be used in multiple senses and hence the mere presence of an explicit cue in a sentence does not imply that it functions as a negator, (e.g., \emph{He could \underline{ not} help me more}). \citet{Reitan2015} reported that in the twitter corpus the cue word \emph{ \underline{no}}  often occurs as an exclamation leading to erroneous predictions.  Such cues should be marked as false negations. 
		\item Negation cues are not part of the scope.
	\end{itemize}
	\item Annotating the Scope
	\begin{itemize}
		\item Annotate  the minimal span for scope.
		\item Scope is continuous.
		\item A noun or an adjective negated in a noun phrase: If only the noun or adjective is being negated then do not annotate the entire clause. Consider each term separately, (e.g., \emph {There are \underline {no} {\bf details} on the return page)}. 
		\item A verb or an adverb phrase: By and large, the entire phrase is annotated, (e.g., \emph{ I do  \underline{ not} {\bf want to update it anymore} }).
	\end{itemize}
\end{itemize}		

 We used  a different scheme for annotating nouns and adverbs as compared to \cite{Councill2010}. Our nouns have a more restricted scope  contrary to the previous work where typically the entire phrase is negated in a noun phrase.

\section{Negation Cue and Scope Detection Experiments}
\label{neg_model}

We divided the dataset  into train and test sets giving a training set of 2317 cues and test set of 604 cues to train both a cue detection and BiLSTM scope prediction.
\subsection{Negation Cue Detection}
The task of cue detection system  is to determine if the potential cue word negates a concept in the sentence. It is based on the state-of-the-art cue classifier described by \cite{Read2012UIO1,Velldal11b,EngVelOvr17}.  A binary SVM classifier is used to disambiguate the cue for only the known cue words, considering the set of cue words as a closed class. Our baseline system uses the features and implementation as described in \citet{EngVelOvr17}. The features used are the word form, POS and lemma of the token, and  lemmas  for previous and next position.  Adding simple features such as position of the cue word in the sentence, POS bigrams  improves the F-score of false negation from a {\bf 0.61 } baseline to  {\bf 0.68} on a test set containing 47 false  and 557 actual negation cues. See Table~\ref{cue-class-res}. 
\begin{table}[!htb]
	\begin{center}
		\begin{scriptsize}
			\begin{tabular} { p{1.3cm} p{1.3cm}  p{1.1cm} p{0.9cm}}  
				\toprule
				&\multicolumn{2}{c}{\bf F-Score}& \\
				\cmidrule(r){2-4}
				&	{ \bf Baseline} & { \bf Proposed }  & { \bf Support} \\ 
				\midrule
				False cues & 0.61  &0.68 &47  \\ 	
				Actual  cues & 0.97 & 0.98 &557 \\ 
				\bottomrule
			\end{tabular}
		\end{scriptsize}
		\caption{\label{cue-class-res}Cue classification on the test set. }
	\end{center}
\end{table}

\subsection{Negation Scope Detection}

 Syntactic structure of the sentence has been often used to determine the scope of negation  using supervised classifiers \cite{Morante2009,Councill2010,Reitan2015,Read2012UIO1,Carrillo2012ucm1}.  Our work is inspired by the previous rule-based approach using constituency tree \cite{Read2012UIO1,Carrillo2012ucm1,Velldal2012}. We build on that work by adding rules based on semantic information, the position of the negation cue in the tree and the projection of its parent based on phrase structure. The syntax tree is obtained using Stanford CoreNLP \cite{Corenlp}.
It is possible that the negation marker may be present in the main clause but semantically belong to the embedded clause. \cite{gotti2008english}  mention that semantic content of copula verbs 
 is subsidiary to that of subject complement, (e.g., \emph {A drunken worker does not become rich}, the negation marker "not" negates the subject complement "rich" rather than the copula verb "become"). Neg-raising is a linguistic phenomenon where certain predicates such as \emph{think, believe} and \emph{seem} occur in the main clause but may be interpreted to negate the complement clause \cite{Fillmore63,Horn1989}. 
We move ahead in a linear order on either finding a copula verb or neg-raising predicates (NRPs). Table ~\ref{linking-verb} contains the list of such verbs used. 
At this point, the algorithm branches based on POS tag of the token. We traverse the  tree in an upward direction until we find a parent with the desired phrase tag as determined by the POS tag of the token. This method differs from the  previous work that finds the first common ancestor enclosing the negation cue and the word immediately after it, and assumes all descendant leaf nodes to the right as its scope \cite{Read2012UIO1}. Our  detailed algorithm  for finding the scope is presented in Figure~\ref{scope-algo}. 
\begin{table}[!htb]
	\begin{center}
		\begin{small}
			\begin{tabular} { p{1.0cm}  p{1.0 cm}  p{1.0 cm} p{1.0 cm}  p{1.0 cm}  } 
				\midrule
				think & believe& seem& appear&feel  \\ 
				grow&look & prove& remain&smell \\ 
				sound &become &might & are& am \\ 	
				been& has&were & was& is  \\ 
				\midrule
			\end{tabular}
		\end{small}
		\caption{\label{linking-verb}  Neg-raising predicates (NRP) and copula verb. }
	\end{center}
\end{table}

\begin{figure}[h!tb]
	
	\begin{center}
			\begin{small}
		\setlist{nolistsep}
		\begin{enumerate}
		
			\itemsep0em 
			\item Traverse the tokens in linear order and stop on finding any cue from the explicit cue lexicon.
			\item Find the next first occurrence of  noun, verb, adverb, adjective.
			\item If the verb is an instance of copula verb  or neg-raising, move to step2 else go to step4.
			\item Branch depending upon POS tag of the token found in step2.
			\setlist{nolistsep}
			\begin{enumerate}
				\itemsep0em
				\item For  nouns and adjectives: \\
				\begin{itemize}
					\item Traverse the tree in upward direction level by level until you reach an ancestor with a tag of  NP, VP, ADJP, SBAR* or S*  for adjectives. For nouns, stop at  NP, SBAR* or S*. \\
					\item If a PP, VP, ADVP, SQ, SINV or SBAR* is a right child of the ancestor, then remove that child.\\
					\item Get all the descendant leaves as scope.
				\end{itemize}
				
				\item For verbs and adverbs:
				\begin{itemize}
					\item Traverse the tree in upward direction level by level until you reach an ancestor  with a tag of VP, SBAR* or S*.\\
					\item If there exists an SBAR*, SQ, or SINV tag as a right child of the ancestor then remove that child.\\
					\item  Get all the descendant leaves as scope.
				\end{itemize}
				
			\end{enumerate}         
			\item  Apply post-processing rules to align the scopes.    
		\end{enumerate}  
\end{small}       
	\end{center}

	\caption{Negation scope detection .
		\label{scope-algo}}
\end{figure}

Though we use SBAR* tags from syntax tree to determine the clause boundaries but it cannot detect all boundaries. We therefore also used  explicit discourse connectives that signal a contrast  relation, or a coordination to limit the scope. These  connectives  act as a boundary for an idea expressed in one clause. For example,
\emph{ To be honest I am not angry but upset},
the scope of \underline {not }as per the rules given in Figure~\ref{scope-algo}, would be  {\bf angry but upset}. Once we find this scope, we use the discourse connective \emph{`but' } to delimit the scope. The list of connectives used is given in Table~\ref{contrast}. \citet{SEM2012SharedMorante} reported that for the SemEval shared task on negation scope detection, most of the systems were post processed to improve their performance. \citet{Read2012UIO1} formulated  a set of slackening heuristics by removing certain constituents at the beginning or end of a scope. This improved the alignment of scopes from an initial 52.42\%  to 86.13\%. Following a similar approach, the post-processing rules were designed  and are given in Figure~\ref{post-rules}.

\begin{figure}[h!tb]
	\begin{center}
		\setlist{nolistsep}
		\begin{itemize}
			\begin{small}
				\item If the scope contains a connective from the prune-connective list  then delimit the scope before the connective.
				\item If the scope contains a  punctuation then delimit the scope  before  the  punctuation marker.
				\item Remove the negation cue from the scope.
				\item Remove the scope words before the cue word, if any.  
				\item If no scope is found after using these rules then predict a default scope as all the tokens up to the first noun, adjective or verb.
				\item  Include the tokens after the negation cue, upto the beginning of the predicted scope.
			\end{small}
		\end{itemize}  
		
	\end{center}
	\caption{ Post-processing heuristic rules.
		\label{post-rules}}
\end{figure}

\begin{table}[!htb]
	\begin{center}
		\begin{scriptsize}
			\begin{tabular} {  p{1.0 cm}  p{1.0 cm} p{1.0 cm}  p{1.0 cm}   p{1.0 cm}} 
				\midrule
				because     & while & until   & however & what\\ 
				but    & though   & although    & nothing & nowhere \\
				whenever& \& & and  & nonetheless & whereas \\ 
				whose   & why & where &wherever&     \\ 
				\midrule
			\end{tabular}
		\end{scriptsize}
		\caption{\label{contrast} Prune-connective list }
	\end{center}
\end{table}

\subsection{Negation Scope Detection Evaluation}
The algorithm is evaluated using two different measures; \emph{token-level} and \emph{scope-level}. Every token can be either in-scope or out of scope. We report the  F-score for both in-scope and out-of-scope tokens. Since the output is a sequence, F-score metrics may be insufficient as it just considers individual tokens. We also report percentage of correct scopes (PCS). Results are given in Table~\ref{tab-nsdresults}.
\begin{table}[!htb]
	\begin{center}
		\begin{small}
			\begin{tabular} { p{1.8cm}  p{1.6cm}  p{1.5 cm} p{1.2cm}}  
				\toprule
				& \bf Punctuation &  \bf  BiLSTM &  \bf  Proposed   \\ 
				\cmidrule(l){2-4}
			
				In-scope (F)  &0.66 &0.88 &0.85   \\ 
				Out-scope (F) &0.87 & 0.97& 0.97 \\
				PCS &0.52& 0.72& 0.72 \\

				\bottomrule	
			\end{tabular}
		\end{small}
		\caption{\label{tab-nsdresults} Negation classifier performance for scope detection with gold cues and scope.}
	\end{center}
\end{table}
The  out-of-scope token  has a higher F-score as compared to in-scope. This is expected since scope tokens are restricted and less in number as compared to out-of-scope (See Table~\ref{tweet-dist}). The in-scope F-score is more  important for the downstream task of sentiment as we apply negation on predicted in-scope tokens for sentiment. The results show that our proposed model is comparable to the BiLSTM model for sentences with gold cues that have an annotated scope, but our model does not require annotated data. For BiLSTM, we used the implementation provided by the authors \footnote{\url {https://github.com/ffancellu/NegNN}}. We also implemented a punctuation  model that marks
as negated all terms from a negation cue
to the next punctuation. ~\citet{FancelluEACL2017}
mentioned punctuation alone as a strong predictor
for negation scope detection task for a majority previous 
of negation corpora. Notably, it performs poorly on our data as our scope is more restricted.  We next show that having a restricted scope  is beneficial to the antonym based negation sentiment prediction.
\section{Sentiment  with Negation Detection  Pipeline }
\label{sentiment}
Here we show the integration of predicted negation scope in  sentiment prediction pipeline. 
We begin with an overview of the data preprocessing, features and modeling, followed by our experimental setup and results. Finally, a comparison of the prediction performance of different systems is presented.
\subsection{Experimental Method }
From our tweet collection, we discarded tweets containing images and Non-English characters and anonymized all user and company handles, giving a dataset of 21746 tweets.
A sentiment annotation task was run on a data annotation platform. Each tweet was initially annotated by 5 annotators using a 4 point (0 to 3) Likert scale~ \cite{likert1932technique} indicating \texttt{Not-At-All}, \texttt{Slight}, \texttt{Moderate} and \texttt{Very} about their perception on the sentiment for a given tweet. We used a set of gold standard questions to filter out the bad annotators, computed the average score for each label, and assigned the maximum score. A tweet is assigned a sentiment label if the maximum score for that label is greater than 1 else it is discarded from the study, giving a labeled dataset of 17779 tweets. 
To compute the inter-annotator agreement, first we measured what percentage of the annotators out of 5 contributed to the final sentiment label and then took the average over all the tweets giving a  78.8$\%$ inter-annotator agreement. 


\subsubsection {Data Pre-processing}
A cleaning module  is incorporated to reduce sparsity when generating word-based features.
We replaced all links/URL by a keyword URL, removed $\#$ from the hashtags, replaced all $@mentions$, and  replaced emojis and emoticons with the word explanation. An entity recognition module  is run to replace the identified entity using a keyword "ENT".
\subsubsection{Features}
TFIDF-based unigram features. \\
Existence of consecutive question and exclamation marks and capitalized words.\\
Emotion lexicon features: A count of the number of words in each of the 8 emotion classes from the NRC emotion lexicon~\cite{mohammad2010emotions} \\
Sentiment lexicons used:\\
\textit{Bing Liu's Opinion Lexicon}~\cite{ding2008holistic};
\textit{The MPQA Subjectivity Lexicon}~\cite{Wilson2005MPQA};
\textit{Sentiment140 Lexicon}~\cite{kiritchenko2014sentiment};
\textit{NRC Hashtag Sentiment Lexicon}~ \cite{kiritchenko2014sentiment}; \\
For a given tweet, we computed minimum, maximum, average and summation of positive and negative scores of the words in the tweet that lies within a negation scope, and the  average sentiment score of the last word in the tweet.\\
Negation handling:  Append "NOT\_" to each word in the scope. 
\subsubsection{SVM Evaluation}

Libsvm \cite{Chang2011} is used to implement the SVM classifier. The tweet annotated dataset was divided into a train and test set (see Table~\ref{tab-sentresults} for the distribution). The training set was further split into a ratio of 85:15 for the validation set. The three parameters w1, w2 and C were tuned using the validation set. The variables w1 and w2 are the penalty associated to a class and C is the regularization. 
 Table~\ref{tab-sentresults} shows the evaluation metric using Precision, Recall and F measure for each class in the test set.

We do not find a major difference for SVMs( w/o negation). This is in spite of using the standard features such as prefixing the tokens in scope with a keyword \emph{NOT\_} and changing the polarity of the sentiment-bearing words using sentiment lexicons as described in previous work.  A possible reason is that  customer service domain is more negative as compared to general review domain See Section ~\ref{disc} for detailed analysis.

\subsection{Neural Network  Evaluation}

{\bf Baseline 1}: Our first baseline  is a single layer CNN as used in ~\cite{Kim14}.  The model  consists of a 1D convolution layer of window size 2 and 64 different filters. The convolution layer takes as input the GloVe embeddings. Max pooling layer is used to reduce the output dimensionality but keep the most salient information.   \\
{\bf Baseline 2}: \cite{WangJL16} presented  a jointed CNN and LSTM architecture.  The features generated from convolution and pooling operation can be viewed as local features similar to ngrams but cannot handle long term dpendencies.  LSTM can handle CNN's limitation by preserving historical information for a long period of time.  Using this as a motivation, we included a convolutional layer
and max pooling layer before the input is fed into an LSTM. A bidirectional LSTM layer is stacked on the convolutions layer and the tweet representation is taken to the fully connected network. \\
{\bf Proposed Negation + Antonym CNN-LSTM }:
We modified the sentence representation learned by replacing a word in the negation scope with it's antonym. Using antonyms  would reduce the Out-of-Vocabulary words as compared to  prefixing a word with "NOT\_" for learning word representations. Replacing all the words upto punctuation with antonyms could entirely change the sentence meaning and hence this required a more restricted and accurate scope detection. We get the predicted scopes from the scope detection model  described in Section~\ref{neg_model}. The antonym list is obtained from  AntNET~\cite{ANTNET17}

For the NN-based approaches, 20\%  data is used for validation and we save the model weights only if the validation accuracy improves.  The outputs of the LSTM are fed through a sigmoid layer for binary classification.  Regularization is performed by using a drop-out rate of 0.2 in the drop-out layer. The model is optimized using the ~\cite{KingmaB14} optimizer. The deep network was implemented using the Keras package~\cite{chollet2015keras}. Hyper-parameter  optimization for the neural network is performed using Hyperas, a python package, based on hyperopt~\cite{Hyperopt}. Results in Table~\ref{tab-sentresults}  show that the antonym based learned representations are more useful for sentiment task as compared to prefixing with NOT\_.  The proposed CNN-LSTM-Our-neg-Ant improves upon the simple CNN-LSTM-w/o neg. baseline with F1 scores improving from 0.72 to 0.78 for positive sentiment and from 0.83 to 0.87 for negative sentiment. Hence Negation coupled with antonyms improves the  sentiment prediction for a customer service domain. 
\begin{table}[!htb]
	\begin{center}
		\begin{small}
			\begin{tabular} {  p{3.5cm}  p{1.0 cm}  p{1.0cm}  p{1.0cm} }  
				\toprule
				&\multicolumn{3}{c } {Positive Sentiment}  \\  
				\cmidrule(l){2-4}
				
				Classifier &Precision & Recall& Fscore   \\ 
				SVM-w/o neg. & 0.57 & 0.72 &0.64\\ 
				SVM-Punct. neg.  &0.58&0.70&0.63 \\	
				SVM-our-neg. & 0.58 & 0.73 &0.65  \\ 
				CNN &0.63&0.83&0.72\\  
				CNN-LSTM &0.71&0.72&0.72\\
				CNN-LSTM-Our-neg-Ant&{\bf0.78}&{\bf0.77}&{\bf0.78}\\

				\midrule
				
				&\multicolumn{3}{c } {Negative Sentiment} \\  
				\cmidrule(l){2-4}
				&Precision & Recall& Fscore   \\ 
				SVM-w/o neg. & 0.78 & 0.86 &0.82  \\
				SVM-Punct. neg. & 0.78  &0.87&0.83 \\ 	
				SVM-Our neg. & 0.80 & 0.87 &0.83  \\ 
				CNN&0.88&0.72&0.79\\
				CNN-LSTM.&0.83&0.83&0.83\\
				CNN-LSTM-our-neg-Ant&{\bf0.87}&{\bf0.87}&{\bf0.87}\\ \\
				
				\midrule\\
				& Train &&  Test \\
				Positive tweets  & 5121&&1320\\
				Negative tweets & 9094&&2244\\

				
				\bottomrule
			\end{tabular}     
		\end{small}
		\caption{\label{tab-sentresults}Sentiment classification evaluation,  using different classifiers on the test set. }
	\end{center}
\end{table}

\section{Discussion}
\label{disc}
In this section, we aim to show the particularities of our dataset, suggesting the reasons why  negation detection did not improve the performance of the lexicon-based SVM when previous work had seen huge performance gains, and intuitions on how the antonym based method gives improvement.
\begin{itemize}
	\item Class Distribution\\
	Our customer service dataset has a much larger number of negative tweets while the benchmark sentiment dataset used in most of the previous systems has positive class as the majority class \cite{kiritchenko2014sentiment,Reitan2015,nakov2013eveval,Mohammadsem13,Tangsem14}. 
	Additionally, \citet{Reitan2015} reported that the classifier struggles with negative class prediction. A F-measure of 0.533 and 0.323 is reported by \citet{Reitan2015} and \citet{Councill2010} respectively, on negative class prediction. In contrast, our baseline classifier achieves a much higher F score of 0.82 on the negative class. 
	\item Cue Word Distribution.\\
	The conversation negation corpus is annotated for both actual negation cues and sentiment. To see if there exists some correlation between the number of cues and sentiment, we calculated the percentage of positive and negative tweets with more than one cue. 19\% of positive tweets contain more than one negation cue while for the negative class it is 48\%. Though we need more evidence to support, but it is possible that the number of negation cues in these conversations  is a strong indicator of negative class, hence the SVM based classifier had better prediction on negative class detection.
	
	\item Sarcasm and Irony\\
	Results in Table~\ref{tab-sentresults} show that the classifier struggles with positive class precision. A sentiment study on user-generated content by \cite{Carvalho2009,Carvalho} has similar class distribution and results to ours. The sentences expressing negative opinions is almost the double of those expressing positive opinions and  the precision of identifying negative opinions ($\approx$ 89\%) is significantly higher than the precision of identifying positive opinions ($\approx$60\%). They confirm the relevance of irony for sentiment analysis by an error analysis of their present classifier stating that a large proportion of misclassifications ($\approx$35\%) derive from their system's inability to account for irony. We then performed some manual error analysis on the incorrect positive predictions for SVM and observed that some of the incorrect predictions were actually sarcastic, see Table~\ref{sarc-tweet}.  \\
	To get an insight on how our method improves these types of predictions, consider the example in Row2 in Table~\ref{sarc-tweet}. It was predicted as positive by SVM, CNN and LSTM due to the positive word "Awesome". 
	Our method detects negation cue word {\bf "not"}  with  {\bf "able" } in it's scope. The antonym dictionary  then is used to replace {\bf "able"} with  {\bf "incapable"}.  Having a strong negative word corrects the prediction  to negative. These results indicate that there is room for improvement for the positive  class but negation handling may not be enough.  A combination of negation and sarcasm may be a useful direction to explore  in future for customer service conversations.
	
	\begin{table}[!htb]
		\begin{center}
			\begin{small}
				\begin{tabular} { p{0.4cm}  p{6.0cm} }  
					\toprule
					\bf {S.No}& \bf  {Tweet}   \\ 
					\midrule
					1&Hi @username - Love u. I'd recommend not displaying the early bird button in the app if it's broken and not working\\
					2&looks like I won't be able to vote because the train is running late. Awesome \\
					
					\bottomrule
				\end{tabular}
			\end{small}
			\caption{\label{sarc-tweet} Negative sarcastic examples.}
		\end{center}
	\end{table}
	
\end{itemize}

\section{Conclusion and Future  Work}
\label{conc}

This paper presented an approach to negation cue and scope detection in customer service interactions 
on Twitter and the impact of using this component for sentiment detection. We refined the annotation 
guidelines for scope representation, gathering a dataset of 2000 labeled tweets. Our  rule based approach based on  syntactic constituents does not require annotated scope data for training, but performs  comparable to state of the art BiLSTM. To evaluate the effectiveness of negation modeling on sentiment detection, we performed experiments using both an SVM and CNN-LSTM  Architecture. There was no significant improvement between the two lexicon based SVMs (with/without negation handling).  The proposed antonym based negation for CNN-LSTM outperformed both a CNN and a combination CNN-LSTM that did not handle negation. The result and error analysis shows that customer service interactions have higher frequency of negation cues, are more skewed towards negative class, and are sometimes sarcastic.  In future,  we plan to study other language phenomenon such as sarcasm in combination with negation.
%

\bibliographystyle{aclnatbib}
\bibliography{negationtweets.bib}
\end{document}